# Automatic Routing System for Intelligent Warehouses

K. T. Vivaldini; J. P. M. Galdames *Member, IEEE*; T. B. Pasqual, R. M. Sobral; R. C. Araújo, M. Becker *Member, IEEE*; and G. A. P. Caurin, *Member, IEEE*

*Abstract*— Automation of logistic processes is essential to improve productivity and reduce costs. In this context, intelligent warehouses are becoming a key to logistic systems thanks to their ability of optimizing transportation tasks and, consequently, reducing costs. This paper initially presents briefly routing systems applied on intelligent warehouses. Then, we present the approach used to develop our router system. This router system is able to solve traffic jams and collisions, generate conflict-free and optimized paths before sending the final paths to the robotic forklifts. It also verifies the progress of all tasks. When a problem occurs, the router system can change the tasks priorities, routes, etc. in order to avoid new conflicts. In the routing simulations each vehicle executes its tasks starting from a predefined initial pose, moving to the desired position. Our algorithm is based on Dijkstra's shortest-path and the time window approaches and it was implemented in C language. Computer simulation tests were used to validate the algorithm efficiency under different working conditions. Several simulations were carried out using the Player/Stage Simulator to test the algorithms. Thanks to the simulations, we could solve many faults and refine the algorithms before embedding them in real robots.

## I. INTRODUCTION

THE routing task may be understood as the process of simultaneously selecting appropriate paths for the AGVs (Automated Vehicle Guided) among different solutions. One of the goals of routing for AGVs is the minimization of cargo cost. In recent years, several algorithms, distinguished in two categories: static [1]-[7] and dynamic [8]-[17], have been proposed to solve routing problems. In the first case, static routing the route from node *i* to node *j* is determined in advance and is always used if a load has to be transported from *i* to *j*. Thus, a simple assumption is to choose the route with the shortest distance from *i* to *j*. However, these static algorithms can not adapt to changes in the system and traffic conditions. In dynamic routing, the information necessary to determine efficient routes are dynamically revealed to the decision-maker and, as a result, various routes between *i* and *j* can be chosen [18]. As one can notice, the several papers found in the literature, have not take into account important features in the routing, such as path maneuvers, time curves, load and unload operations, etc.

Manuscript received February 28, 2010. This work was supported in part by FAPESP (Processes 2008/10477-0, 2008/09755-6, and 2009/01542-6), CAPES (Process 33002045011P8) and CNPq (Process 134214/2009-9).

K. T. Vivaldini; J. P. M. Galdames; T. B. Pasqual, R. M. Sobral; R.C. Araújo, M. Becker; and G. A. P. Caurin are with the Mechatronics Laboratory - Mechanical Engineering Department, at EESC-USP, SP 13566-900 Brazil (e-mails: {kteixeira}{galdames}{becker}{gcaurin} @sc.usp.br; and {thales.bueno}{rafael.sobral}{roberto.carlos.araujo} @usp.br).

In many researches, the route is calculated considering the minimum path. Broadbent et al. [1] presented the first concept of conflict-free and shortest-time AGV routing. The routing procedure described employs Dijkstra's shortest path algorithm to generate a matrix, which describes the path occupation time of vehicles.

Kim and Tanchoco [9] proposed an algorithm based on Dijkstra's shortest-path method and the concept of time windows graph for dynamic routing of AGVs in a bidirectional path network. They presented formulas to consider the time curves, but this was not considered in their results.

Maza and Castagna [12-13] added a layer of real time control to the method proposed by Kim and Tanchoco. In [12], they proposed a robust predictive method of routing without conflicts, and in [13] they developed two algorithms to control the AGV system using a predictive method when the system is subject to risks and contingences avoiding conflicts in real time manner. In both, the curve time is not mentioned.

Möhring et al. [14] extended the approaches of Huang, Palekar and Kapoor [18] and Kim and Tanchoco [9], and presented an algorithm for the problem of routing AGVs without conflicts at the time of route computation. In the preprocessing step the real-time computation for each request consists in the determination of the shortest path with time-windows and a following readjustment of these time-windows, both is done in polynomial time. By goal-oriented search, they computed times appropriate for real-time routing. Extending this concept, Klimm et al. [14] presented an efficient algorithm to cope with the problem of congestion and detours that also avoids potential deadlock situations. The authors in [14-15] did not consider that the time window for routes with curves could imply the estimated time. Only in [16], Gawrilow et al. presented an algorithm to avoid collisions, deadlocks and livelocks already at the route computation time, considering the physical dimensions of the AGVs to cope with complex maneuvers and certain safety aspects that imply particular applications.

Ravizza [17] focused on the online control of AGVs, presenting a heuristic approach for task assignment and a dynamic polynomial-time sequential routing to guarantee deadlock and conflict-free routing of the AGVs in undisturbed operations. The articles cited above used Dijkstra's Algorithm to calculate the route.

However, the shortest route is not always the most efficient method, ie. the number of maneuvers is larger than the number necessary for the path to be executed.

In this context, as a solution to this problem, this paper presents the development of a routing system that computes optimal routes, reducing the amount of unnecessary maneuvers, allowing path planning and coordinating the task execution of the transport and handling in structured environments. It also avoids known and dynamic obstacles located in the environment. This approach investigates an efficient solution to routing AGVs, as an alternative to the various methods developed previously in the literature. We also propose a software architecture that considers the local and global tasks of the robotic forklifts (e.g.: local navigation, obstacle avoidance, and auto-localization). Figure 1 presents this architecture.

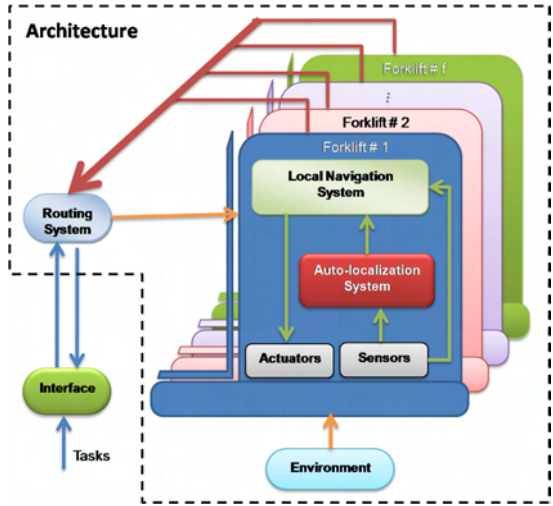

Fig. 1. Overall System Architecture proposed.[25]

In summary, it can be explained as follows: the routing system receives information about the required transportation tasks from a user interface. Based on these data, it selects the minimum quantity of robotic forklifts necessary to execute the tasks. Then, based on a topologic map of the environment, it calculates the routes for the selected forklifts, checking possible collisions, traffic jams, etc. After that, it sends the routes to each robotic forklift and regularly verifies the progress of all tasks. Taking into account the global route, each forklift calculates its own local path necessary to reach its goal and monitors its surroundings looking for mobile or unexpected obstacles during the execution of the planned path. At a certain frequency, based on sensor data and an environment map, each robotic forklift auto-localization subsystem updates its estimated position and informs these data to the local navigation subsystems. The local navigation subsystem compares the current position with the desired one. If the robot deviates from the route, the local navigation subsystem sends commands to correct its pose returning the planned route. If the local navigation subsystems verify that the route has exceeded the limit runtime determined by the routing, it communicates it to the routing system, which recalculates the route avoiding collisions or deadlocks among other robotic forklifts. The algorithm is based on Dijkstra's shortest-path method [20] to calculate the routes of the robotic forklift adding heuristic functions to optimize the quantity of maneuvers, and using the method of routing with time-windows [21] to ensure conflict-free routes.

## II. ROUTING ALGORITHM

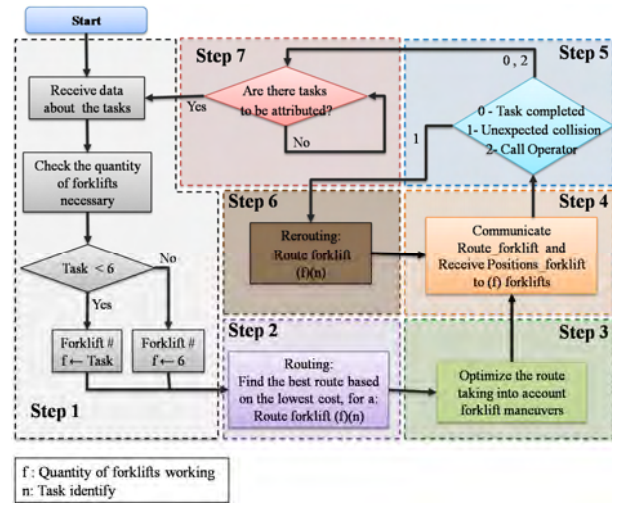

Fig. 2. Proposed Algorithm.

The routing algorithm (Fig. 2) was based on the dynamic programming approach, which consists in dividing the original problem into smaller and simpler problems. This approach was very useful to our problem, as it presents a sequence of decisions to be taken along a time sequence [22]. Therefore, our routing system could be divided into seven steps.

### A. Step 1

The routing system receives a list of the requests as input (Table 1 presents a summary of the requested data considered in the storage activities). Each request is a task defined by a sequence of pairs: loading stations (origin node) and unloading stations (destination node). Then, according to the requests, the routing system checks the quantity of robotic forklifts necessary to execute the tasks, and assigns each task to a forklift.

TABLE I
REQUEST DATA, INPUTS AND OUTPUTS OF OUR ALGORITHM

| Data | Input | Output |
|---|---|---|
| *Request* (Orders) | Quantities, loading data / unloading data of each product | Necessary number of forklifts and allocation of each request in a route for the forklifts |
| *Loading* | Location point for loading the pallets | Routes that the forklifts should execute to carry the pallets |
| *Unloading* | Location point for delivering the pallets | Routes that the forklifts should execute to unload the pallets |

Several tasks can be attributed to various robotic forklifts, but one task cannot be attributed to several robotic forklifts. When all tasks have been designated and the quantity of robotic forklifts necessary has been determined, these data are sent to Step 2 in order to calculate the routes.

*B. Step 2*

In this Step the routing system applies a graph-based approach. The graph is obtained by using a topological map of the warehouse environment. The route necessary to execute each task is composed of two sub-routes. Which their one has its own origin and destination nodes. Then Dijkstra's Algorithm is applied to calculate the lowest path (relation between distance and total cost) for each robotic forklift. The path is a continuous sequence of positions and orientations of the robotic forklift (e.g.: the intermediate positions and the pre-established positions and orientations present in Fig. 3). Aiming to guarantee that collisions between robotic forklifts will not occur, the forklift cannot use the same arc or node at the same time. Therefore, we know when every arc is (or not) occupied by a robotic forklift for each calculated sub-route. These data (time intervals $[a_i, b_i]$) and the time window are saved on a list (log file), allowing verifying possible conflicts between paths of the robotic forklifts. When the sub-route is a task, a value representing the time needed to load or unload the pallets is added to the destination node time interval.

*C. Step 3*

In this step the quantity of maneuvers (e.g.: curves) during the path route is analyzed by heuristic functions, which verify the possibility of optimizing them. If the maneuvers are unnecessary, the route is optimized using again the Dijkstra's Algorithm with time-window, taking into account the costs previously established. If the resulting total execution time of the task is longer than the previously calculated (Step 2), the final route is not changed (it means that the route cannot be optimized). At the end of this step, if the route has been optimized, it will be conflict-free.

*D. Step 4*

In this step, the routes are sent to each robotic forklift and the routing system interacts with the other systems embedded in the robots (navigation and auto-localization systems), as shown in Fig. 4.

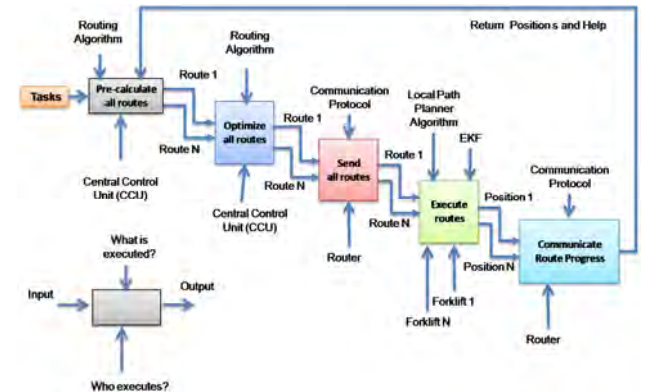

Fig. 4. Interaction of the routing system with navigation, control, and localizations systems.

The routing system verifies each task progress regularly. Each robotic forklift has its own sensor, auto-localization, and local navigation sub-system running independently and informing regularly the global path planner about the status of each task (position and problem found or task finished). Therefore, it is possible to minimize the impact of this local problem in the global system that controls the intelligent warehouse. Table 2 presents a summary of this procedure.

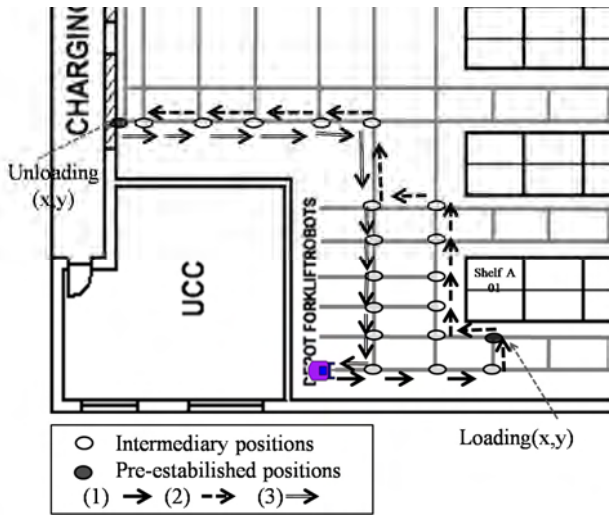

Fig. 3. Illustration of route definition. A robotic forklift will execute a task (go to Shelf A- 01, load a pallet and unload it at Charging Platform F). After this task, it will return to its depot. The first (1) and the second (2) sub-routes represent the task execution. The third one (3) represents the robotic forklift returning to its depot.

Figure 3 illustrates how algorithm generally defines a route. The origin of the first sub-route is the current robotic forklift position. Its destination is defined as the task initial position (loading node). The second sub-route origin is the loading node and its destination is the unloading node. If the same robotic forklift has another task, the following sub-routes are defined in a similar way. If it is closing-time, or the robotic forklift needs to recharge its batteries, after the last task its final sub-route drives it directly to its depot. Thus, we always have the origin and destination nodes and also know if a task is (or not) in execution.

TABLE II
REQUEST DATA, INPUTS AND OUTPUTS OF OUR ALGORITHM

| Data | Input | Output |
|---|---|---|
| Problems in route execution | Route cannot be concluded. | Inform position and list of the problems found (unexpected collisions, exceeded time) or call operator. |

*E. Step 5*

In this step the algorithm checks the status of the tasks, where: 0 - task finished, 1 - rerouting of the task (unexpected collisions or exceed time), and 2 - call operator and assign tasks of this forklift to others.

If the status is equal 1 to, it is necessary to recalculate the sub-route (Here, the use of sub-routes is very useful as, only the sub-route is corrupted). Basically the algorithm verifies which robotic forklift presents conflicts and its location (arc or node – see Fig. 5). The reroute process is carried out in Step 6.

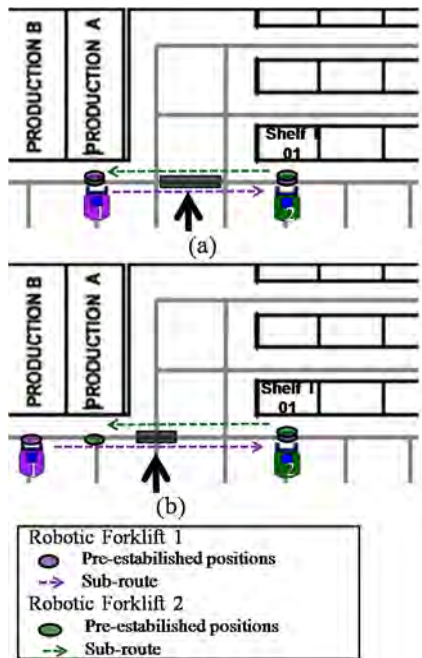

Fig. 5. Illustration of two examples of collision between the robotic forklifts #1 and #2. In (a) a collision in the arc, in (b) in the node.

*F. Step 6*

Now, the corrupted sub-route is eliminated and the arc where the conflict occurred is blocked. Then, the sub-routes of the robotic forklifts that caused the conflict are recalculated (Fig. 6).

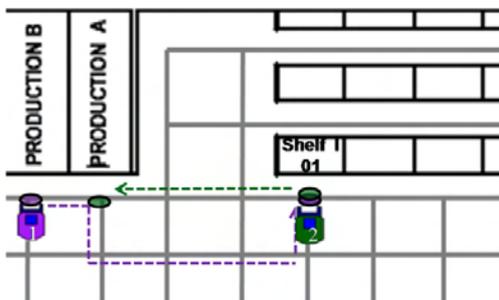

Fig. 6. Result of the rerouting process for the example presented in Fig. 5-b.

To recalculate this sub-route again, the algorithm returns to steps 2 and 3, the origin node receive the position where forklift is and the destination node is the same for this sub-task. These new sub-routes are called conflict-free sub-routes. If after these conflict-free sub-routes there are other sub-routes in the forklift schedule, then it is necessary to verify the conflict-free final time in the destination node. If it differs from the previously calculated (Step 2), then it is necessary to readjust the time windows of this forklift checking if it may cause other conflicts. For instance, let us assume that the robotic forklift #1 in Fig. 5 has more sub-routes scheduled, after its conflict-free sub-route (Fig. 7).

Therefore, it is necessary to readjust the time windows of the subsequent sub-routes by checking the list of available time windows. Then, the algorithm verifies again the presence of conflicts in the route, due to the readjustment of the time windows. If there is any conflict, it returns to the beginning of this step.

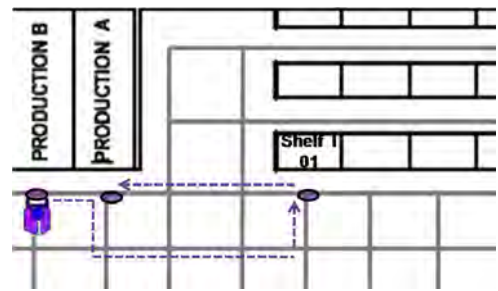

Fig. 7. Sub-routes of the robotic forklift #1.

*G. Step 7*

Finally, the algorithm verifies if there are more tasks to be executed. If so, it returns to its beginning (step 1). The objective of this step is to verify and validate the execution time of the tasks attributed versus position of the robotic forklifts.

### III. INTERFACE

In this work, a graphic interface allows the routing operation to be controlled by an operator, through Qt Creator 4 software [23] and a simulated virtual environment in the Player/Stage developed in previous works [24]. Both were implemented in Linux Operating System.

The graphic interface (Fig. 8) was developed to perform the communication between the user and the routing system (including all system algorithms). It also allowed the configuration of both routing system and simulation environment at a higher level (Fig. 9), guaranteeing the functionality of the routing operation and returning information concerning the process performance in a simple and visual way.

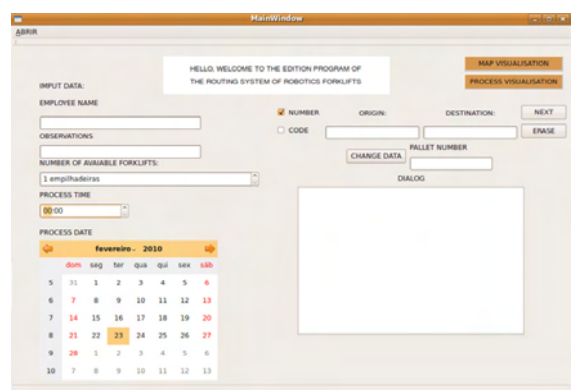

Fig. 8. Graphic Interface

The interface allows the operator to schedule processes of robotic forklifts, setting time, data, quantity of pallets involved, as well as origin and destination, and the quantity of mobilized forklifts. With such information, the operator can simulate the process. In order to do this, it is possible to configure the simulation environment in the most appropriate and realistic way, providing drivers for the available sensors and the bi-dimensional plan of the modeled warehouse. Thus, the operator can simulate and evaluate with a good accuracy the time and effectiveness of the process without the mobilization of real robotic forklifts.

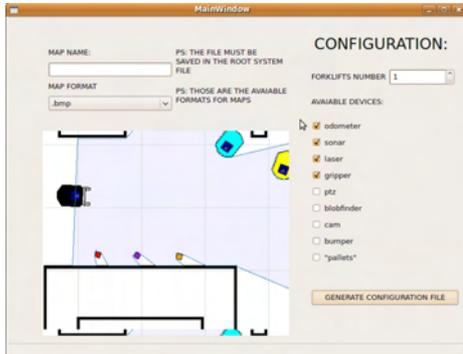

Fig. 9. Graphic Interface for the simulated virtual environment.

## IV. RESULTS

As previously mentioned, this work used a local navigation and auto-localization systems in order to obtain more realistic simulations. The routing system uses the A* algorithm in the local path planning, and applies the Extended Kalman Filter (EKF) algorithm for the auto-localization for each robotic forklift [24][25]. The proposed algorithm is based on Dijkstra's shortest-path method to calculate the routes of the robotic forklift adding heuristic functions to optimize the quantity of maneuvers, and using the method of routing with time-windows to ensure conflict-free routes. Using heuristic functions, it was possible to verify the number of unnecessary curves and make a route optimization reducing the amount of maneuvers to be performed and thus increasing the mobility of the robotic forklift. It is important to emphasize that the optimized route do not exceeds the total cost previously established.

In order to verify the performance of our algorithms, several simulations were carried out using a virtual warehouse (50x30m) and the Player/Stage Simulator. In the simulation 6 robotic forklifts working at the same time in this warehouse were considered (at this moment we did not consider here unknown obstacles like people walking or other vehicles moving in the warehouse). In order to simplify the implementation, we selected in the Player/Stage library the Pionner mobile robot to represent our robotic forklifts. Maximum cruiser speeds of 1m/s and 5°/s were applied to all robots. The algorithm could solve the conflicts and return the optimized routes. Figure10 shows the simulations of the tasks attributed for 6 robotic forklifts. This test would have several collisions that can be observed in Fig. 10-a (arrow). Firstly, the deadlock between the robotic forklifts #4 (red) and #5 (yellow) is detected. Due to this deadlock, a new deadlock also involving the robotic forklift #6 (lemon) occurs. The collisions happened because the robotic forklift #4 needs to occupy the same arc at the same time that robotic forklifts #5 and #6. Unfortunately, as we highlighted before, if the Router did not verify all sub-routes after the first corrupted one detected, more conflicts may occur. In this case, a new deadlock between the robotic forklifts #6 (lemon) and #1 (purple). Figure 10-b presents the proposed algorithm. One may verify that the conflicts identified in Fig. 10-a were eliminated, the final paths provided by the routing algorithm were collision-free and avoided traffic jams. It also shows in detail the optimizations performed by the Routing System.

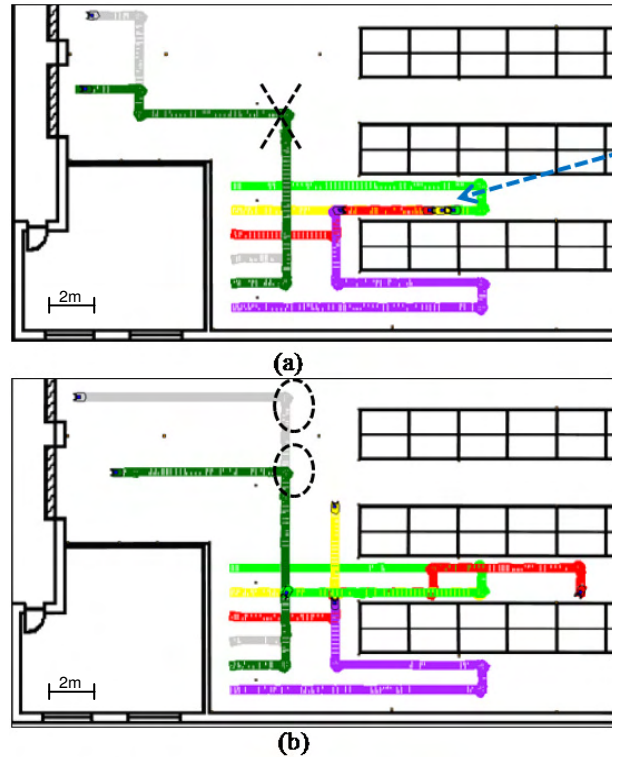

Fig. 10. Comparison between shortest path method and optimized routes of the robotic forklift #3 (grey) and forklift #2(green). In (a) the route marked with an X represents the maneuver that the robotic forklifts perform using shortest path method; and in (b) the routes marked with circles represent the maneuvers that the robotic forklifts did using the route optimized by our algorithm.

In this scenario, each robotic forklift calculates the trajectory to be executed considering the route sent by the routing system (Fig. 11). Each forklift receives a safety time for the minimum path between two nodes, them navigation system can fix both the speed and trajectory to execute the path. The local planner applies time windows to calculate the safe trajectory. In case of disturbances that divert the robot from its planned trajectory, the local planner tries to run the route within the stipulated time, e.g., increasing the speed. If the time is exceeded, the robot sends an error log to the routing system describing the problem found during the route.

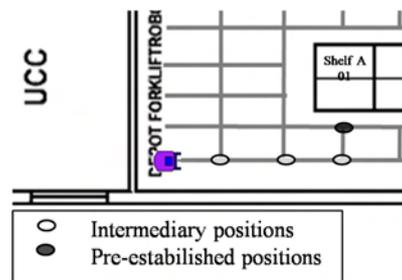

Fig. 11. Illustration of route informed for local planner. The routing system sends the each two points (e. g. forklift position and Intermediary position) the safety time that the local planner has to calculate the path planning.

V. CONCLUSIONS

Firstly, we described the most important works found in literature that focus on routing systems for AGVs. Next, we presented the approach used to develop our Routing system. It is able to solve traffic jams and collisions. It also guarantees optimized routes before sending the final paths to the robotic forklifts. It also verifies the progress of all tasks. When a problem occurs, the routing system can change the tasks priorities, routes, etc. to avoid new conflicts. In order to verify the algorithm's performance, the robotic forklifts were tested executing tasks that simulate the load and unload of goods using the interface and the virtual environment in the Player/Stage software. The use of the interface was very interesting because it allowed us to control the process simulation at a higher level. Therefore, this program can be used directly in industries at future.

The A* algorithm was a simple way to control the robot maneuvers, and the EKF was sufficient to avoid the position error propagation. Both algorithms were implemented in C. As a result, we could solve many faults and refine the algorithms before embedding them in real robots. We also plan some improvements on the algorithm, for instance, to refine the routing task planner and to take into account more realistic parameters during the simulations (e.g.: velocity changes while maneuvering, load and unload times, etc.). It is important to emphasize that in the current version, the algorithm assigns the time for each path according to the distance, maneuvers and loading and unloading of pallets. The robotic forklift speed during the path is determined by the local planner, which checks the distance to achieve the goal and determines the speed required for the robotic forklift to arrive there in time. One limitation of the algorithm is that collisions are always solved by finding a new route. A possible solution would be the application of stoppages to one of the forklifts during a certain time interval, or the reduction of its speed. Both options will be investigated in future. At this moment our group is finishing the design and construction of mini-mobile robots and a 1:5 scale warehouse to test experimentally our algorithms.